\documentclass{article}
\usepackage{times,url,graphicx,amssymb,caption}

\newcommand{\bcX}{{\bf X}}
\newcommand{\bcY}{{\bf Y}} 
\newcommand{\bff}{{\bf f}}

\newcommand{\bx}{{\bf x}}
\newcommand{\by}{{\bf y}}

\newtheorem{Theorem}{Theorem}

\newtheorem{Definition}{Definition}

\newtheorem{Postulate}{Postulate}
\newtheorem{Corollary}{Corollary}

\newenvironment{proof}{Proof:}{$\blacksquare$}

\date{11 February 2014}

\title{Justifying Information-Geometric Causal Inference}

\author{
Dominik Janzing, Bastian Steudel, Naji Shajarisales, Bernhard Sch\"olkopf\thanks{\texttt{first.last@tuebingen.mpg.de}}  \\
Max Planck Institute for Intelligent Systems     \\
Spemannstr. 38\\
72076 T\"ubingen\\
Germany\\
}

\begin{document}

\maketitle

\begin{abstract}
Information Geometric Causal Inference (IGCI) is a new approach to
distinguish between cause and effect for two variables.
It is based on an independence assumption between input distribution and
causal mechanism that can be phrased in terms of 
orthogonality in information space.
We describe two intuitive reinterpretations of this approach that makes 
IGCI more accessible to a broader audience.

Moreover, we show that the described independence is related to the hypothesis
that unsupervised learning and semi-supervised learning only works for predicting the cause from the effect and not vice versa.
\end{abstract}

\section{Information-Geometric Causal Inference}

\label{sec:int}

While conventional causal inference methods \cite{Spirtes1993,Pearl2000} use conditional independences 
to infer a directed  acyclic graph of causal relations among at least three random variables, there is a whole family of recent methods that
employ more information  from the joint distribution than just conditional independences \cite{Zhang_UAI,Hoyer,DiscrAN,deterministic,Info-Geometry,Jonas_tpami,UAI_identifiability}. 
Therefore, these methods can even be used for inferring the causal relation between just two observed variables  (i.e., the task to infer
whether $X$ causes $Y$ or $Y$ causes $X$, given that there is no common cause and exactly one of the alternatives is true, becomes solvable).  

As theoretical basis for such inference rules, 
\cite{Algorithmic,LemeireJ2012} postulate the following asymmetry between cause and effect: if  $X$ causes $Y$ then $P(X)$ and $P(Y|X)$ 
represent independent mechanisms of nature and therefore contain
no information about each other. Here, ``information'' is understood in the sense of description length, i.e.,
knowing $P(X)$ provides no shorter description of $P(Y|X)$ and vice versa, if description length is
identified with Kolmogorov complexity. This makes the criterion empirically undecidable because Kolmogorov complexity is uncomputable \cite{Vitanyi08}.
 \cite{anticausal} pointed out that ``information'' can also be understood
in terms of predictability and used this to formulate the following hypothesis:
semi-supervised learning (SSL) is only possible from the effect to the cause but not visa versa. This is because, if $X$ causes $Y$,
knowing the distribution $P(Y)$ may help in better predicting $X$ from $Y$ since it  
may contain information about $P(X|Y)$, but $P(X)$ cannot help in
better predicting $Y$ from $X$.  

Information-Geometric Causal Inference (IGCI) \cite{deterministic,Info-Geometry}  has been proposed
for inferring the causal direction between just two variables $X$ and $Y$. 
In its original formulation it applies only to the case where $X$ and $Y$ are related by an invertible functional relation, i.e.,
$Y=f(X)$ and $X=f^{-1}(Y)$, but some positive empirical results have also been reported for noisy relations \cite{deterministic,Info-Geometry}.
We will also restrict our attention to the noiseless case. This is because attempts to generalize the theory to non-deterministic 
relations are only preliminary \cite{Info-Geometry}. Moreover,  the deterministic toy model nicely shows  
 {\it what kind} of dependences between $P(Y)$ and $P(X|Y)$ occur while the dependences in the non-deterministic case are not yet well understood.

We first rephrase how
IGCI has been introduced in the literature and then  explain our new interpretations. They also provide a better intuition about the relation to SSL. For didactic reasons we restrict the
 attention to the case where $f$ is a monotonously increasing diffeomorphism of $[0,1]$.
We assume that $P(X)$ and $P(Y)$ have strictly positive densities $p_X$ and $p_Y$. We often write $p(x)$ instead of $p_X(x)$ whenever this causes no confusion.
Then \cite{deterministic} assumes:

\begin{Postulate}[uncorrelateness between density and log slope] \label{psot:uncorr}
For $X$ causing $Y$,
\begin{equation}\label{uncorr}
\int_0^1 p(x) \log f'(x) dx  = \int_0^1  \log f'(x) dx \quad  \hbox{ (approximately) }\,.
\end{equation}
\end{Postulate}

The interpretation that (\ref{uncorr}) is an independence condition becomes more clear when the functions $x \mapsto p(x)$ and $x\mapsto \log f'(x)$
are interpreted as random variables on $[0,1]$. Then, the difference  
 between the left and the right hand side of (\ref{uncorr}) 
is the covariance of $p_X$ and $\log f'$ with respect to the uniform distribution 
 \cite{Info-Geometry}. 
The intuition is
that it is unlikely, if $f$ and $P(X)$ are chosen independently, 
that regions where the slope of $f$ is large (i.e. large $\log f'$),
meet regions where $p_X$ is large and others where $p_X$ is small.

\begin{figure}
\centerline{
\includegraphics[width=0.3\textwidth]{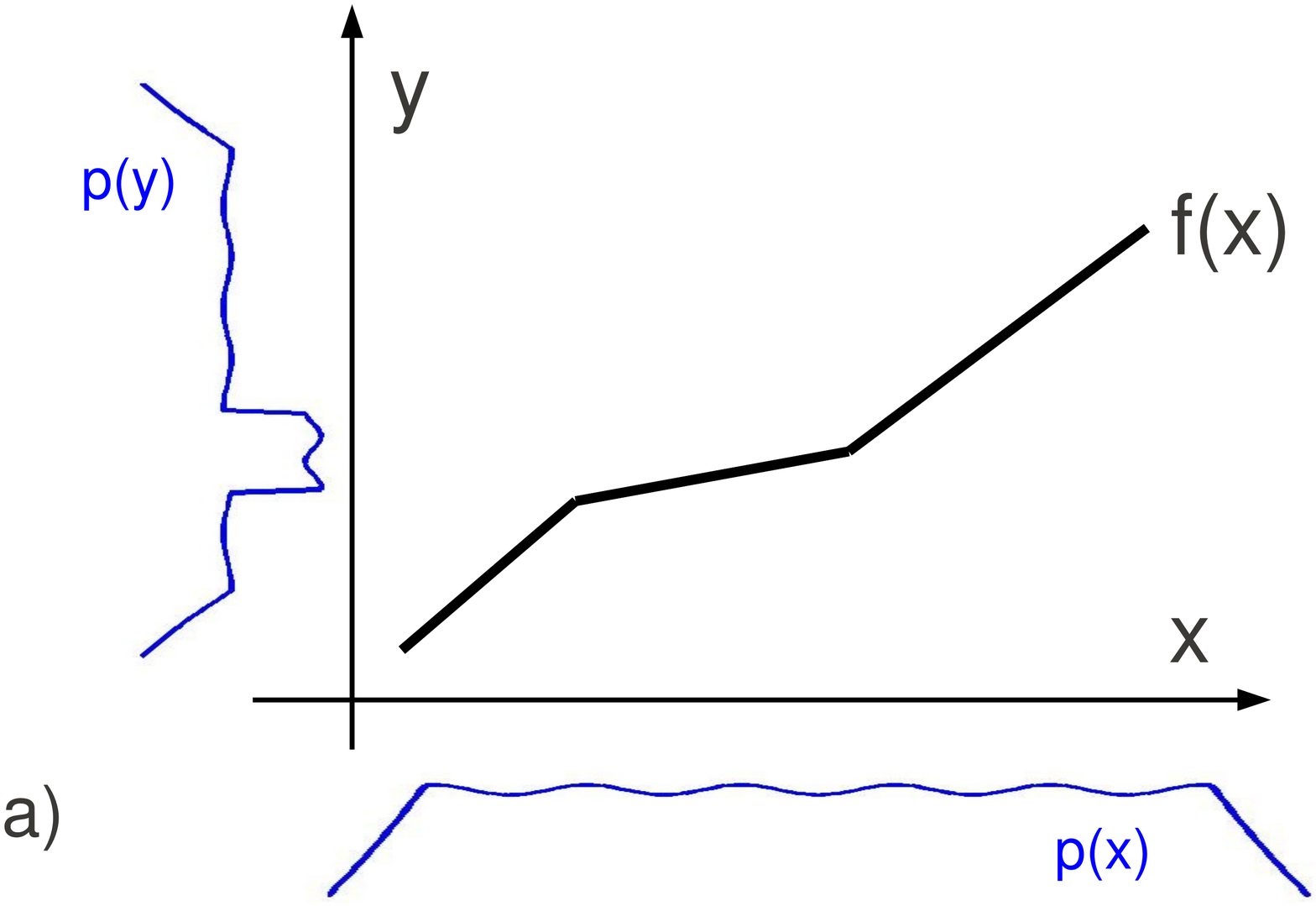}
\hspace{3cm}
\includegraphics[width=0.33\textwidth]{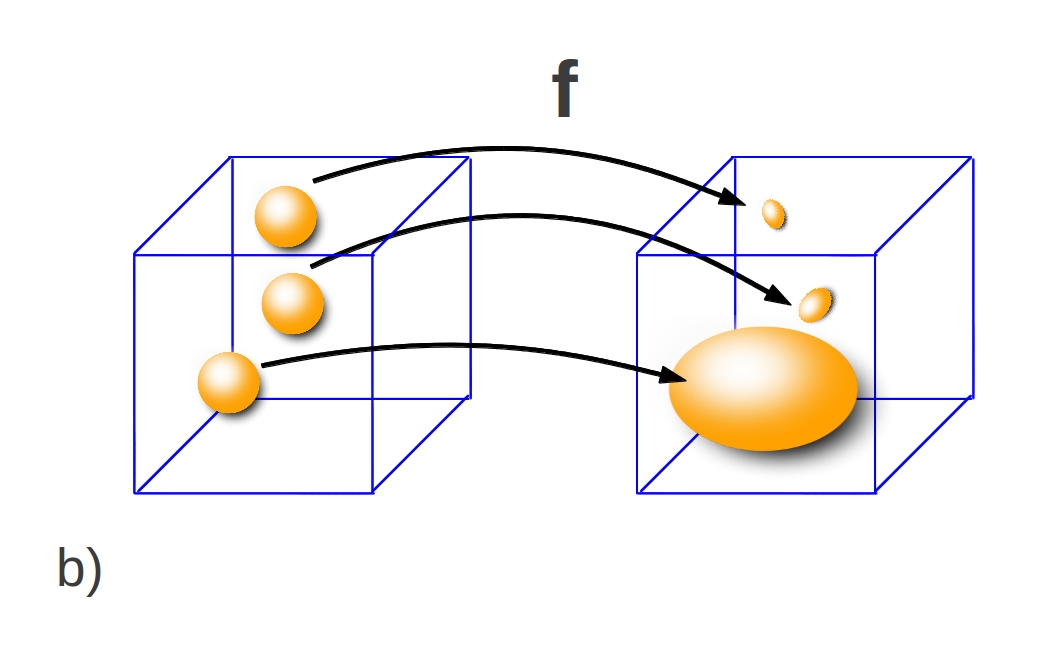}
}
\caption{\label{fig:corr} a) (taken from \cite{Info-Geometry}) if the fluctuations of $p_X$ don't correlate with the slope of the functions, regions of high density $p_Y$ tend to occur 
more often in regions where $f$ is flat. b) 
 $\bff$ maps the cube to itself. The regions of points $\bx$ with large $|\det \bff'(\bx)|$ (here: the leftmost sphere) can only be a small fraction of the cube.}
\end{figure}

Simple calculations  \cite{deterministic,Info-Geometry} show that (\ref{uncorr}) implies that $p_Y$ is positively correlated with the slope of $f^{-1}$ since
\begin{equation}\label{pc}
\int_0^1 p(y) \log f^{-1'}(y) dy \geq \int_0^1  \log f^{-1'}(y)  dy\,,
\end{equation}
with equality iff $f'\equiv 1$. 
This  is illustrated in figure~\ref{fig:corr} a).

Moreover, using $\int \log f'(x)dx \leq \log \int f'(x)dx =0$, eq.~(\ref{uncorr}) implies 
\begin{equation}\label{expec}
0\geq \int_0^1 p(x) \log f'(x) dx =: C_{X\rightarrow Y}\,.
\end{equation}
Using
\begin{equation}\label{minus}
C_{X\rightarrow Y} = - \int_0^1 p(y) \log f^{-1'} (y) dy=:C_{Y\rightarrow X}\,,
\end{equation}
we get
\begin{equation}\label{infmeth}
C_{X\rightarrow Y}  \leq C_{Y\rightarrow X}\,,
\end{equation}
with equality only for $f\equiv id$.
For empirical data $(x_1,y_1), \dots, (x_n,y_n)$,
with $x_1<x_2\cdots <x_n$ (and hence $y_1 < y_2 \cdots < y_n$),
this suggests the following inference method:

\begin{Definition}[Information-Geometric Causal Inference]\label{def:IGCI}
Infer $X\rightarrow Y$ whenever
\begin{equation}\label{empIGCI}
\sum_{	j=1}^{n-1} \log \frac{|y_{j+1} -y_j|}{|x_{j+1}-x_j|}
< \sum_{j=1}^{n-1} \log \frac{|x_{j+1} -x_j|}{|y_{j+1}-y_j|}\,.
\end{equation}
\end{Definition}

Some robustness of IGCI with respect to adding noise has been reported \cite{Info-Geometry} when
the following modification is used: on the left hand side of (\ref{empIGCI})
the $n$-tuples are ordered such that $x_1<x_2\cdots <x_n$, while
the right hand side assumes $y_1<y_2<\cdots <y_n$. Note that in the noisy case, the two conditions are not equivalent. Moreover, the left hand side of (\ref{empIGCI}) is no longer minus the right hand side since eq.~(\ref{minus})  no longer makes sense. Albeit hard to formulate explicitly, it is intuitive to consider the
left hand side as measuring ``non-smoothness'' of $P(Y|X)$ and the right hand side the one of $P(X|Y)$. Then, the causal direction is the one with the smoother conditional.

To describe the information theoretic content of (\ref{uncorr}) and (\ref{pc}), we introduce the uniform distributions $u_X$ and $u_Y$
for $X$ and $Y$, respectively. Their images under $f$ and $f^{-1}$ are given gy
the probability densities $\overrightarrow{p}^{f}_Y (y)=f^{-1'}(y)$ and $\overleftarrow{p}^{f^{-1}}_X(x)=f'(x)$, respectively.
We will drop the superscripts $f$ and $f^{-1}$ whenever the functions they refer to are clear. 
 Then
 (\ref{uncorr}) reads 
\begin{equation}\label{uncorrdens}
\int p(x) \log \frac{\overleftarrow{p}(x)}{u(x)} dx = \int u(x) \log \frac{\overleftarrow{p}(x)}{u(x)} dx\,,
\end{equation}
and
is equivalent to  the following additivity of relative entropies \cite{deterministic,Info-Geometry}:
\begin{equation}\label{orth1}
D(p_X\|\overleftarrow{p}_X) = D(p_X\| u_X) + D(u_X\| \overleftarrow{p}_X)\,.
\end{equation}
Likewise, (\ref{pc}) reads
\begin{equation}\label{pcdens}
\int p(y) \log \frac{\overrightarrow{p}(y)}{u(y)} dy \geq \int u(y) \log  \frac{\overrightarrow{p}(y)}{u(y)} dy\,.
\end{equation}
In the terminology of Information geometry \cite{Amari}, (\ref{orth1}) means that
the vector connecting $p_X$ and $u_X$ is orthogonal to the one connecting $u_X$ and $\overrightarrow{p}_X$. Thus
the ``independence'' between $p_X$ and $f$ has 
been phrased in terms of orthogonality, where $f$ is represented by $\overleftarrow{p}_X$. 
Likewise, the dependence between $p_Y$ and $f^{-1}$ corresponds to the fact that the vector connecting $p_Y$ and $u_Y$ 
is not orthogonal to the one connecting $u_Y$ and $\overrightarrow{p}_Y$.  
The information-theoretic formulation motivates why one postulates uncorrelatedness of $p_X$ and $\log f'$ instead of
one between $p_X$ and $f'$ itself. A further advantage of this reformulation is that 
 $u_X$ and $u_Y$ can then be replaced with other ``reference measures'', 
 e.g., 
Gaussians with the same variance and mean as $p_X$ and $p_Y$, respectively (which is more appropriate for variables with unbounded domains)
\cite{deterministic,Info-Geometry}.

However, both conditions (\ref{uncorr}) and (\ref{orth1}) are quite abstract. 
Therefore, we want to approach IGCI from completely different directions.
In section~\ref{sec:untyp}  we will argue that a large positive value for $C_{X\rightarrow Y}$ shows
that the observed $n$-tuple $(x_1,\dots,x_n)$ is untypical in the space of all possible $n$-tuples.
In section~\ref{sec:numb} we show that condition (\ref{empIGCI}) implies that there are, in a sense, more functions from $X$ to $Y$ than vice versa.
In section~\ref{sec:ssl} we  explain why the correlation between distribution and slope that occurs in the anticausal direction helps for
unsupervised and semi-supervised learning.

\section{First reinterpretation: untypical points}

\label{sec:untyp}

\begin{figure}
\centerline{
\includegraphics[width=0.3\textwidth]{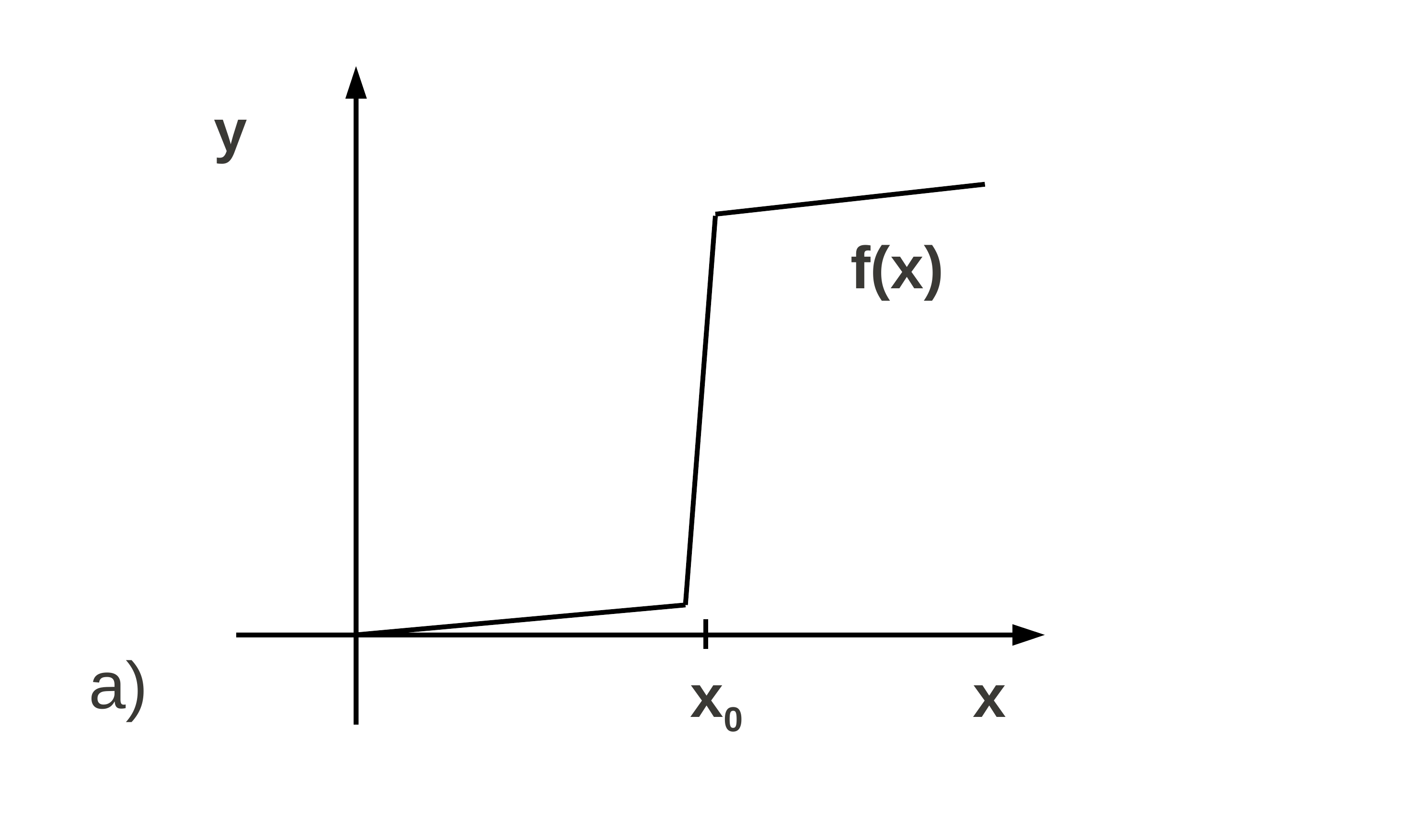}
\includegraphics[width=0.3\textwidth]{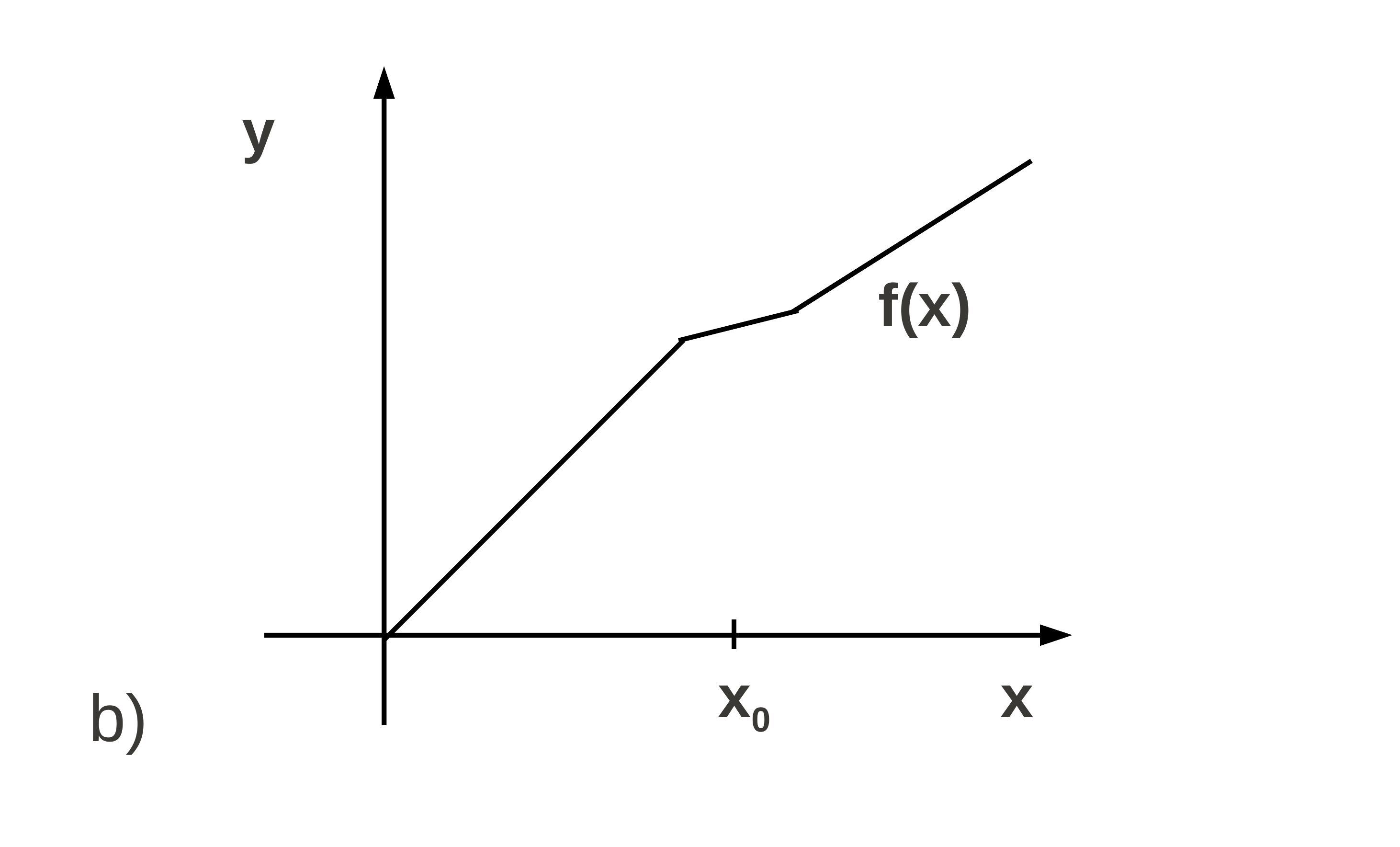}
\includegraphics[width=0.3\textwidth]{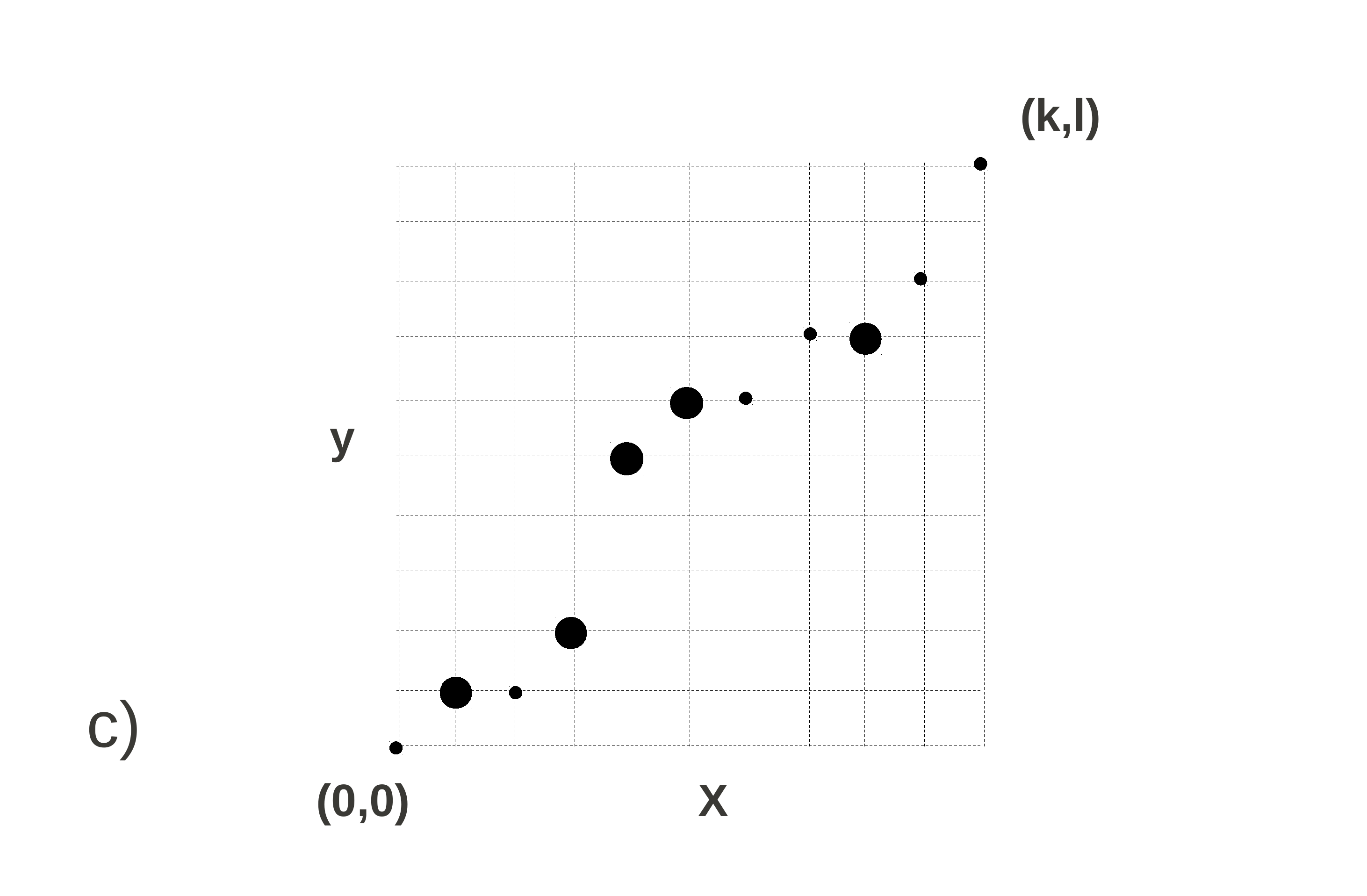}
}
\caption{\label{fig:one_point} (a) $x_0$ has an untypical position in both cases because it meets in a small region with large slope 
or in a region with small slope (b). c) Function on a grid. The large dots denote the observed points, the small ones visualize one option of
interpolating by a monotonic function that crosses $(0,0)$ and $(k,\ell)$.}
\end{figure}

Let us consider again a monotonously increasing diffeomorphism $f:[0,1]\rightarrow [0,1]$ and explain
in which sense a point $x\in [0,1]$ can have a ``typical'' or an ``untypical''  position relative to $f$.
Consider the function $f$ shown in Figure~\ref{fig:one_point} a). The point $x_0$ is untypical because it meets $f$ in a region whose slope 
is larger than for the majority of points. Of course, $x_0$ can also be untypical in the sense
that the slope of $x_0$ is smaller than for the majority of points, see Figure~\ref{fig:one_point} b).
There is, however, an important asymmetry between large slope and small slope: if the slope at $x_0$ is significantly higher than
the {\it average} slope over the entire domain, then  $x_0$ is necessarily untypical because the slope can significantly exceed the average only
for a small fraction of points. If the slope is significantly {\it below} the average, this does not mean that 
the point is untypical because this may even be the case for most of the points, as one can easily see on Figure~\ref{fig:one_point} a).
This asymmetry is known from statistics: a non-negative random variable may quite often attain values that are smaller than their expected value by orders of magnitude, but exceeding the expectation by a large factor is unlikely due to the Markov inequality.

The above idea straightforwardly generalizes to mappings between multi-dimensional spaces: then a point ${\bf x}$ can be untypical relative to a function $\bff$ in the
sense that the Jacobian of $\bff$ is significantly larger than the average. This is, for instance, the case for the points in the 
leftmost sphere of Figure~\ref{fig:corr} b).
We first introduce ``untypical'' within a general measure theoretic setting:

\begin{Theorem}[untypical points]\label{thm:unt}
Let $U(X)$ and $U(Y)$ be probability distributions on
measure spaces $\bcX$ and $\bcY$, respectively. Let
$\bff:\bcX \rightarrow \bcY$ be measurable and let the image of $U(X)$ under $\bff$ have a strictly positive density $q_\bff$ with respect to $U(Y)$.
Then, points $\bx$ for which $-\log q_\bff (\bff(\bx)) \gg 0$ 
are unlikely (``untypical'') in the sense that 
\begin{equation}\label{mkin}
U_X \left\{ \bx\in \bcX \,| -\log q_\bff (\bff(\bx)) \geq c \right\} \leq e^{-c}\,,
\end{equation}
for all  $c>0$.
\end{Theorem}

\proof{ Let $U_X^\bff$ be the image of $U_X$ under $\bff$. Then
the left hand side of (\ref{mkin}) can be rephrased as
\begin{equation}\label{image}
U^\bff_X \left\{ \by\in \bcY \,| -\log q_\bff (\by) \geq c \right\}=
U^\bff_X \left\{ \by \in \bcY \,|\, \frac{1}{q_\bff} (\by) \geq e^{c} \right\}\,.
\end{equation}
Note that $\int 1/q_\bff \, dU_X^\bff=\int dU_Y= 1$, therefore 
$1/q_\bff$ is a non-negative random variable with expectation $1$
and the right hand side  of (\ref{image}) is smaller than $e^{-c}$ by the Markov inequality.
$\blacksquare$
}

\begin{Corollary}[diffeomorphism between hypercubes]\label{cor:hyper}
Let $\bff :[0,1]^n \rightarrow [0,1]^n$ be a diffeomorphism. 
Then the volume of all points $\bx$ for which 
$
\log |\det \bff'(\bx)| \geq c
$
is at most $e^{-c}$.
\end{Corollary}  

\proof{ Follows from 
$
q_\bff (\bff(\bx)) = |\det \bff^{-1'}(\bff(\bx))|=1/|\det \bff'(\bx)|\,,
$ if $U(X), U(Y)$ are the Lebesgue measure. 
$\blacksquare$
}

The corollary is very intuitive: since the average of $|\det \bff'(\bx)|$ 
over the  hypercube is $1$, 
the fraction of $\bx$ for which
the Jacobian is significantly larger than $1$ is small.
%On the other hand, there is no such statement for small Jacobian, the
%fraction of points for which $|\det \bff'(x)|$ is close to zero could
%be arbitrarily close to $1$.
Whenever we observe a point $\bx$ whose Jacobian is significantly larger than $1$, we are skeptical about whether it has been chosen independently of $\bff$. 
%In other words, $\bff$ contains information about $\bx$ because knowing $\bff$ allows for a shorter description of $\bx$.

We now describe in which sense IGCI rejects observations $\bx:=(x_1,\dots,x_n)$  that are untypical.
For
$
\bff(\bx):=(f(x_1),\dots,f(x_n))
$
we observe
\begin{equation}\label{fact}
\log |\det \bff'(\bx)| =\sum_{j=1}^{n} \log |f'(x_j)|\,.
\end{equation}
If the right hand side of (\ref{fact}) is significantly larger than zero,
then $\bx$ is untypical because this holds
only for a small fraction of the hypercube. This suggests the following reinterpretation of
IGCI: due to (\ref{expec}) and (\ref{minus}),  the right hand side of (\ref{empIGCI}) will usually be positive when $X\rightarrow Y$
is the true causal direction. Whenever it attains a ``large'' positive value, the 
expression $\sum_{j=1}^n \log f^{-1'} (y)$ is also large because the former is an approximation of the latter. 
Then, $\by:=(y_1,\dots,y_n)$ is untypical for the function $(f^{-1},\dots,f^{-1})$, which makes us rejecting  
the causal hypothesis $Y\rightarrow X$.

\section{Second view: counting the number of functions}

\label{sec:numb}
 
We now argue that IGCI, roughly speaking, 
 amounts to choosing the direction for which there is a larger number of functions that fit the data. 
We will also discuss some connections to inductive principles of statistical learning theory \cite{Vapnik06}.

To get a clear sense of ``number of functions'', we discretize the interval $[0,1]$ for $X$ and $Y$ and
assume that all $(x_j,y_j)$ are taken from the grid
\[
\{0,\dots,k\}\times \{0,\dots,\ell\}\,,
\]
as in Figure~\ref{fig:one_point} c).
We assume furthermore that $0=x_1<x_2<\cdots < x_n=k$ and, similarly, $0=y_1<y_2<\cdots < y_{n}=\ell$
and denote these observations by $\bx:=(x_1,\dots,x_n)$ and $\by:=(y_1,\dots,y_n)$.
Let $F_{\bx,\by}$ be the set of all monotonic functions for which $y_j=f(x_j)$ with $j=1,\dots,n$.
Our decision  which  causal direction is more plausible for the observation $(\bx,\by)$ will now be based on 
the following generating models:
a function $f$ is chosen uniformly at randomly from $F_{(0,k),(0,\ell)}$, i.e., the set of functions from $X$ to $Y$ 
that pass the points $(0,0)$ and $(k,\ell)$. Then, each $x_j$ with $j=2,\dots,n-1$ is chosen uniformly at random from 
$\{0,\dots,k\}$. This yields the following distribution on the set of possible observations: 
\begin{equation}\label{genXY}
P_{X\rightarrow Y}(\bx,\by):=\frac{1}{(k+1)^{n-2}} \frac{|F_{\bx,\by}|}{|F_{(0,k),(0,\ell)}|}\,.
\end{equation}
Likewise, we obtain a distribution for the causal direction $Y\rightarrow X$ given by
\begin{equation}\label{genYX}
P_{Y\rightarrow X}(\bx,\by):=\frac{1}{(\ell+1)^{n-2}} \frac{|G_{\by,\bx}|}{|G_{(0,\ell),(0,k)}|}\,,
\end{equation}
where $G_{\by,\bx}$ denotes the corresponding set of functions from $Y$ to $X$.\\
For a general grid $\{0,\dots,a\}\times \{0,\dots,b\}$, elementary combinatorics shows that 
the number of monotonic functions from $X$ to $Y$
that pass the corners $(0,0)$ and $(a,b)$  is given by
\[
N(a,b) = { a+ b -1\choose b}\,.
\]
Therefore, 
\begin{equation}\label{comb}
\frac{|F_{(0,k),(0,\ell)}|}{|G_{(0,\ell),(0,k)}|} =  \frac{N(k,\ell)}{N(\ell,k)}=\frac{k}{\ell}\,.
\end{equation}
The pair $(\bx,\by)$ defines $n-1$ grids $\{x_j,\dots,x_{j+1}\} \times \{y_j,\dots,y_{j+1}\}$ and  $|F_{\bx,\by}|$  
is the product of the numbers for each grid.
Thus,
\begin{equation}\label{combcomb}
\frac{|F_{\bx,\by}|}{|G_{\by,\bx}|} =\prod_{j=1}^{n-1} \frac{N(x_{j+1}-x_j,y_{j+1}-y_j)}{N(y_{j+1}-y_j,x_{j+1}-x_j)}  
= \prod_{j=1}^{n-1} \frac{x_{j+1} -x_j}{y_{j+1}-y_j} \,,
\end{equation}
where we have applied rule (\ref{comb}) to each grid $\{x_j,\dots,x_{j+1}\}\times \{y_j,\dots,y_{j+1}\}$.  
Combining (\ref{genXY}), (\ref{genYX}), (\ref{comb}), and (\ref{combcomb}) yields
\begin{equation}\label{igcid}
\frac{P_{X\rightarrow Y} (\bx,\by)}{P_{Y \rightarrow X}(\bx,\by)}
=\frac{(l+1)^{n-2}}{(k+1)^{n-2}}\cdot \frac{\ell}{k}\prod_{j=1}^{n-1} \frac{(x_{j+1}-x_j)}{(y_{j+1}-y_j)} 
\,.
\end{equation}
We now consider the limit of arbitrarily fine grid, i.e.,
$k,\ell \to\infty$ (while keeping the ratios of all $x_j$ and those of all $y_j$ constant). Then 
expression (\ref{igcid}) becomes independent of the grid and can be replaced with
\begin{equation}\label{discrIGCI}
 \sum_{j=1}^{n-1} \log  
 \frac{(x_{j+1}-x_j)/(x_n-x_1)}{(y_{j+1}-y_j)/(y_n-y_1)}\,.
\end{equation}
Thus, IGCI as given by Definition~\ref{def:IGCI} simply compares the
loglikelihoods of the data with respect to the two competing  generating models above
since  (\ref{discrIGCI}) coincides with the left hand side of (\ref{empIGCI}) after normalizing $x_j,y_j$ such that $x_n-x_1=y_n-y_1$.

The above link is intriguing, but the function counting argument
required that we discretized the space, leading to finite function
classes, and it is not obvious how the analysis should be done in the
continuous domain. In statistical learning theory \cite{Vapnik06}, the
core of the theoretical analysis is the following: for consistency of
learning, we need uniform convergence of risks over function
classes. For finite classes, uniform convergence follows from the
standard law of large numbers, but for infinite classes, the theory
builds on the idea that whenever these classes are evaluated on finite
samples, they get reduced to finite collections of equivalence classes
consisting of functions taking the same values on the given
sample.\footnote{Strictly speaking, this applies to the case of
  pattern recognition, and it is a little more complex for regression
  estimation.}  In transductive inference as well as in a recently
proposed inference principle referred to as inference with the
``Universum,'' the size of such equivalence classes plays a central
role \cite{Vapnik06}. 

The proposed new view of the IGCI principle may be linked to this
principle.  Universum inference builds on the availability of
additional data that is not from the same distribution as the training
data --- in principle, it might be observed in the future, but we haven't
seen it yet and it may not make sense for the current classification
task.\footnote{E.g., the task might be digit recognition, but the
  Universum points are letters.}  

Let us call two pattern recognition functions equivalent if they take
the same values on the training data. We can measure the size of an
equivalence class by how many possible labellings the functions from
the class can produce on the Universum data. A classifier should then
try to correctly separate the training data using a function from a
large equivalence class --- i.e., a function from a class that allows
many possible labellings on the universum data, i.e., one that does not
make a commitment on these points.

Loosely speaking, the Universum is a way to adjust the geometry of the
space such that it makes sense for the kind of data that might come
up. This is consistent with a paper that linked the Universum-SVM
\cite{Weston06} to a rescaled version of Fisher's discriminant
\cite{Sinz08}. Taken to its extreme, it would advocate the view that
there may not be any natural scaling or embedding of the data, but
data points are only meaningful in how they relate to other data
points.  

In our current setting, if we are given a set of Universum points in
addition to the training set, we use them to provide the
discretization of the space. We consider all functions equivalent that
interpolate our training points, and then determine the size of the
equivalence classes by counting, using the Universum points, how many
such functions there are. The size of these equivalence classes then
determines the causal direction, as described above --- our analysis
works exactly the same no matter whether we have a regular
discretization or a discretization by a set of Universum points.

\section{Employing the dependences for un- and semi-supervised regression}

\label{sec:ssl}

We now argue that the correlations between $p_Y$ and $\log f^{-1'}$ 
are relevant for prediction in two respects:
First, knowing $f^{-1}$ tells us something about $p_Y$, and second, $p_Y$ tells us something about $f^{-1}$.
Note that section~\ref{sec:untyp} already 
describes the first part: assume $C_{Y \rightarrow X}=\sum_{j=1}^n \log f^{-1'}(y_j)$ is large. Then, knowing $f^{-1}$
(and, in addition, a lower bound for $C_{Y\rightarrow X}$) restricts the set of possible  
$n$-tuples to a region with small volume. 

We know explore what $p_Y$ tells us about $f^{-1}$. 
This scenario is the one in
unsupervised and semi-supervised learning (SSL) \cite{SSL} 
since  the distribution
of unlabeled points is used to get information about the labels.
\cite{anticausal} hypothesized that this is not possible in causal direction, i.e., if the labels are the effect. In anticausal direction, 
the labels are the cause and unsupervised learning employs the information that $P({\rm effect})$ contains about $P({\rm  cause}|{\rm effect})$. 
As opposed to the standard notation in machine learning, where
$Y$ is the variable to be predicted from $X$,
 regardless of which of the variables is the cause, we prefer to keep the convention that $X$ causes $Y$ throughout the paper. Thus, we consider the task of predicting $X$ from $Y$ and discuss in which sense knowing the distribution $p_Y$ helps. We study this question within the finite grid
$\{1,\dots,k\}\times \{1,\dots,\ell\}$
 to avoid technical difficulties with defining priors on the set of
 differentiable functions. We use essentially the generating model $X\rightarrow Y$ from section~\ref{sec:numb} with monotonic functions
on the grid with the following modification: we restrict the set of functions to the set of surjective  functions $F_s$ to ensure that
the image of the uniform distribution is a strictly positive distribution. To avoid that this is a strong restriction, we assume that 
$k\gg \ell$. Since we use the grid only to motivate ideas for the continuous case, this does not introduce any artificial asymmetry
between $X$ and $Y$.
Then we assume that a function $f$ is drawn uniformly at random from $F_s$. Afterwards, $n$ $x$-values are drawn uniformly at random from $\{1,\dots,k\}$. 
This generating model defines a joint distribution  
for $n$-tuples $\bx,\by$ and functions $f$ via
\[
P_{X\rightarrow Y} (\bx,f,\by):=\frac{1}{|F| k^{n}} \delta(\by-f^n(\bx))\,,
\] 
where $f^n$ denotes the application of $f$ in each component\footnote{Note that marginalization of $P_{X\rightarrow Y}(\bx,f,\by)$ to $\bx,\by$ yields
the same distribution as in section~\ref{sec:numb} up to the technical modifications of having fixed endpoints and surjective functions.}.

 In analogy to the continuous case, we introduce the image of the uniform distribution 
on $\{1,\dots,k\}$ 
under $f$ by
$\overrightarrow{p}^f(y):=|f^{-1}(y)|/k$ and
obtain
\[
\log P_{X\rightarrow Y}(\by |f)= \sum_{j=1}^{n} \log \overrightarrow{p}^f(y_j) \,.
\]
Hence,
\begin{equation}\label{X>Y}
\log P_{X\rightarrow Y} (f|\by)=  \sum_{j=1}^{n} \log \overrightarrow{p}^f(y_j) +\log \frac{1}{|F|} - \log P_{X\rightarrow Y} (\by) \,,
\end{equation}
where we have used the fact that all functions are equally likely.
We rephrase (\ref{X>Y}) as 
\begin{equation}\label{predf}
\log P_{X \rightarrow Y}(f|\by)= n \sum_y p(y) \log \overrightarrow{p}_f(y) +c(\by)\,,
\end{equation}
 where $p_Y$ denotes the distribution of empirical relative frequencies defined by the $n$-tuple $\by$ 
and $c$ is a summand that does not depend on $f$. (\ref{predf}) provides a prediction of $f$ from
 $\by$. 
We now ask why this
 prediction should be {\it useful} although it is based on the wrong model
 because we assume that the true data generating model does not draw
 $x$-values from the uniform distribution  (instead, $p_X$ only ``behaves
 like the uniform one'' in the sense of (\ref{uncorr})).
 To this end, we show that the likelihood of $f$ is unusually high compared to other functions that are, in a sense, equivalent.
To define a set of equivalent functions, we
first represent $f$ by the following list of non-negative integers:
\[
|f^{-1}(1)|, |f^{-1}(2)|,\dots,|f^{-1}(\ell)|\,,
\]
and observe that this list describes $f$ uniquely because $f$ is monotonic.
Then every permutation $\pi$ on $\{1,\dots,\ell\}$ defines a new monotonic function
$f_\pi$ by the list
$|f^{-1}(\pi(j))|$ with $j=1,\dots,\ell$.
Note that $\overrightarrow{p}^{f_\pi} (y)=\overrightarrow{p}^f(\pi^{-1}(y))$.
Therefore, one can easily argue that for large $\ell$, most permutations induce
functions $f_\pi$ for which
\begin{equation}
\sum_y p(y) \log \overrightarrow{p}^{f_\pi} (y) \approx
\frac{1}{\ell} \sum_y   \log \overrightarrow{p}^{f_\pi} (y)\,.
\end{equation}
This is because the difference between left and right hand side can be interpreted as covariance of
the random variables $p_Y$ and $\log \overrightarrow{p}^f_Y$ with respect to the uniform distribution on $\{1,\dots,\ell\}$ (see also section~\ref{sec:int}) and
a random permutation yields approximately uncorrelated samples with high probability\footnote{Note that more precise statements 
would require lower bounds on $\log \overrightarrow{p}^f$ and upper bounds on $p_Y$, which goes beyond the scope of this paper.}.
Therefore, if we observe that
 \[
 \sum_y p(y) \log \overrightarrow{p}^f(y) 
 > \frac{1}{\ell} \sum_y  \log \overrightarrow{p}^f(y)  \,,
 \]
 in the sense of significant violation of equality, 
 the true function $f$ has a higher likelihood than 
 the overwhelming majority of the functions $f_\pi$. In other words,
 $P_{X\rightarrow Y}(f|\by)$ prefers the true function within a huge set
 of functions that are equivalent in the sense of having the same numbers of pre-images.
  
Translating this into  the continuous setting, we infer $f$ from $p_Y$ by
defining a loglikelihood function over some appropriate set of sufficiently  smooth functions via
\begin{equation}\label{flike}
\log P_{X\rightarrow Y}(f|p_Y):= \mu \int p(y) \log \overrightarrow{p}^f(y)dy  +c(p_Y) \,,
\end{equation}
with a free parameter $\mu$,
since we have explained in which sense this provides a useful
prediction in the discrete setting. 

 Rather than getting a distribution over the possible functions for $f$ we often
want to get a single function $\hat{g}$ that predicts $X$ from $Y$, i.e., 
an estimator for $g:=f^{-1}$. 
We define
\begin{equation}\label{ghat}
\hat{g}(y):=\int_0^y p(y') dy' = \int_0^{f^{-1}(y)} p(x') dx'\,,
\end{equation}
and observe that $\hat{g}$ maps $p_Y$ to the uniform distribution due to $\hat{g}'(y)=p(y)$, i.e.,
$\hat{g}$ provides the correct prediction if $p_X$ is uniform. 
Moreover, its inverse $\hat{g}^{-1}$ is the unique maximizer of (\ref{flike}) since it maps $u_X$ to $p_Y$.

\begin{figure}
\centerline{
\includegraphics[width=1.1\textwidth]{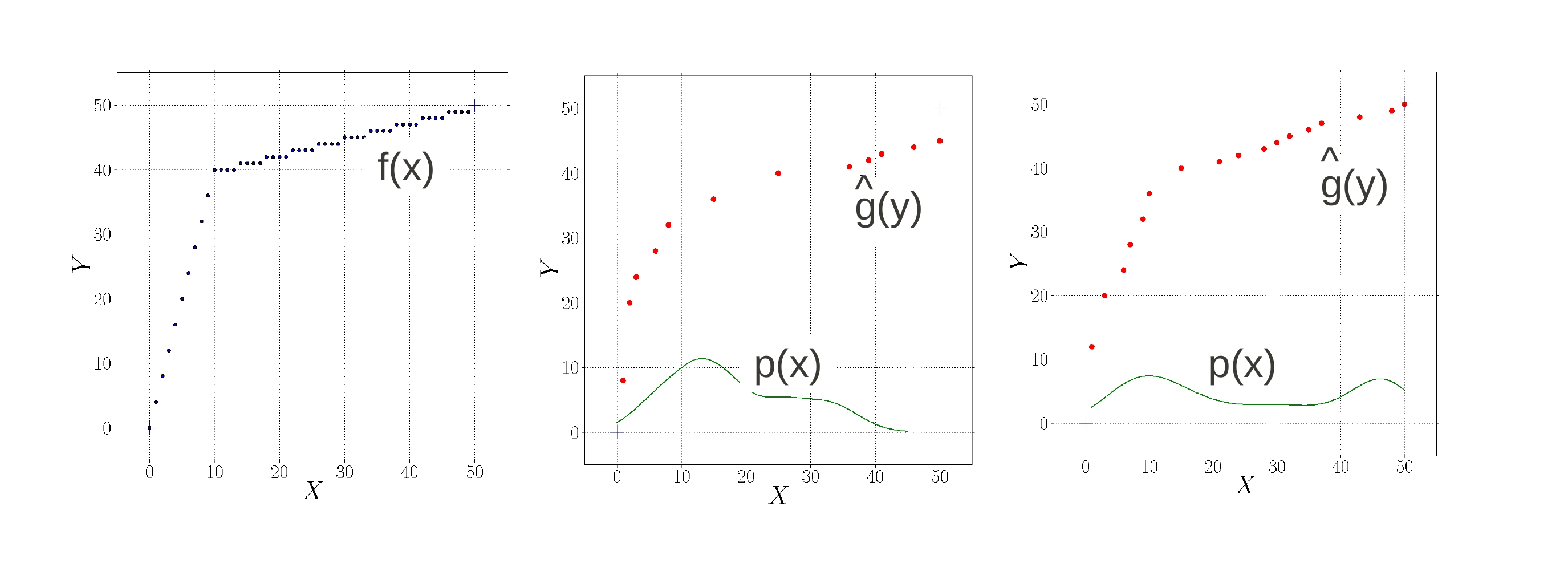}
}
\caption{\label{fig:exp} A simple function  $f$ from $X$ to $Y$ (left) and the functions $\hat{g}$ inferred from the empirical distributions of $Y$ for two different
input distributions $p_X$.}
\end{figure}

To understand in what sense $\hat{g}$ still provides a good prediction even 
if $p_X$ strongly deviates from $u_X$, we observe that
the error remains small if the cumulative distribution function $CDF(x):= \int_0^x 
p(x') dx'$ does not deviate too much from the one for the uniform distribution.
Furthermore, $\hat{g}$ shares some qualitative behavior with $g$ because it tends to have large slope where $g$
has large slope because $\hat{g}'$ correlates with $\log g'$ due to (\ref{pc}).
Figure~\ref{fig:exp} visualizes unsupervised prediction based on $\hat{g}$ for a simple function.

We now argue that information theory provides theoretical results on how close $\hat{g}$ is to $g$.
To this end, we define an (admittedly uncommon) distance of functions by the relative entropy distance of the densities that they map to
the uniform distribution.
 Thus, 
$
D(p_Y\|\overrightarrow{p}_Y)
$
measures the distance between $\hat{g}$ and $g$.
Since relative entropy is conserved under bijections \cite{cover}, we have 
\begin{equation}\label{transf}
D(p_Y\|\overrightarrow{p}_Y)=D(p_X\|u_X)\,,
\end{equation}
i.e., the deviation between $\hat{g}$ and $g$ coincides with the deviation of $p_X$ from the uniform distribution. 
Together with (\ref{orth1}), (\ref{transf}) implies
\begin{equation}\label{bpred}
D(p_Y\|\overrightarrow{p}_Y)\leq D(p_X\|\overleftarrow{p}_X)\,,
\end{equation}
with equality only for $\overleftarrow{p}_X=u_X$. 
Note that $p_X$  represents the functions 
 $\hat{f}$ obtained from the analog of (\ref{ghat}) 
when trying to infer $f$ from $p_X$ (although we know that this is pointless when $p_X$ and $f$ are chosen independently).
Since $\overleftarrow{p}_X$  represents the true function $f$, 
we conclude: no matter how much $\hat{g}$ deviates from $g$, $\hat{f}$ deviates even more from $f$, i.e., the error of 
unsupervised prediction in causal direction  always exceeds the one in anticausal direction.

For the semi-supervised version, we are given a few labeled points $(y_1,x_1),\dots,(y_l,x_l)$ as well as a large number of unlabeled points
$y_{l+1},\dots,y_{n+l}$. We consider again the limit where $n$ is infinite and the observations tell us exactly the distribution $p_Y$. 
Then we use the information that $p_Y$ provides on $f$ for interpolating between the labeled points via
\[
\hat{f}^{-1}(y):=x_j +(x_{j+1}-x_j) \frac{\int_{y_j}^{y} p(y') dy'}{\int_{y_j}^{y_{j+1}} p(y')dy' } \,,
\]
whenever $y\in [y_j,y_{j+1}]$. 

Note that the above schemes for un- and semisupervised prediction are not supposed to compete with existing 
methods for real-world applications (the assumption of a noiseless invertible relation does not occur too often anyway). 
The goal of the above ideas is only to present a toy model that shows that the independence between
$P({\rm cause})$ and $P({\rm effect}|{\rm cause})$ typically yields 
a dependence between $P({\rm effect})$ and $P({\rm cause}|{\rm effect})$ that can be employed for prediction.
Generalizations of these insights to the noisy case could be helpful for practical applications.

\section*{Acknowledgements}

The authors are grateful to Joris Mooij for insightful discussions.

%\bibliographystyle{unsrt}
%\bibliography{../../../SVN_Dominik/literatur/literatur}

\end{document}